\documentclass[final]{cvpr}

\usepackage{times}
\usepackage{epsfig}
\usepackage{graphicx}
\usepackage{amsmath}
\usepackage{amssymb}
\usepackage[ruled,vlined]{algorithm2e} 

\SetCommentSty{mycommfont}

\newcommand{\CHI}{\raisebox{2pt}{$\chi$}}
\newcommand{\Graph}{\mathcal{G}}

\usepackage[pagebackref=true,breaklinks=true,colorlinks,bookmarks=false]{hyperref}

\def\cvprPaperID{9} 
\def\confYear{CVPRW 2021}
\begin{document}

\title{Superpixels and Graph Convolutional Neural Networks for Efficient Detection of Nutrient Deficiency Stress from Aerial Imagery}

\author{Saba Dadsetan\\
University of Pittsburgh\\
Intelinair, Inc.\\
{\tt\small sabadad@cs.pitt.edu}
\and
David Pichler\\
Intelinair, Inc.\\
{\tt\small david.pichler@intelinair.com}
\and
David Wilson\\
Intelinair, Inc.\\
{\tt\small david@intelinair.com}
\and
Naira Hovakimyan\\
University of Illinois at Urbana Champaign\\
Intelinair, Inc.\\
{\tt\small naira@intelinair.com}
\and
Jennifer Hobbs\\
Intelinair, Inc.\\
{\tt\small jennifer@intelinair.com}
}

\maketitle

\begin{abstract}
   Advances in remote sensing technology have led to the capture of massive amounts of data. Increased image resolution, more frequent revisit times, and additional spectral channels have created an explosion in the amount of data that is available to provide analyses and intelligence across domains, including agriculture.  
   However, the processing of this data comes with a cost in terms of computation time and money, both of which must be considered when the goal of an algorithm is to provide real-time intelligence to improve efficiencies. Specifically, we seek to identify nutrient deficient areas from remotely sensed data to alert farmers to regions that require attention; detection of nutrient deficient areas is a key task in precision agriculture as farmers must quickly respond to struggling areas to protect their harvests.
   Past methods have focused on pixel-level classification (i.e. semantic segmentation) of the field to achieve these tasks, often using deep learning models with tens-of-millions of parameters.
   In contrast, we propose a much lighter graph-based method to perform node-based classification. We first use Simple Linear Iterative Cluster (SLIC) to produce superpixels across the field. 
   Then, to perform segmentation across the non-Euclidean domain of superpixels, we leverage a Graph Convolutional Neural Network (GCN). This model has 4-orders-of-magnitude fewer parameters than a CNN model and trains in a matter of minutes.

\end{abstract}

\section{Introduction}

Nutrient monitoring and management is a central task for farmers throughout the growing season.
To prevent major losses, growers must identify and respond to crop deficiencies promptly while avoiding excess applications of fertilizers that are costly and may have detrimental environmental consequences.
Monitoring the health of crops, potentially across hundreds of thousands of acres of fields spread over many miles, is a daunting task without the aid of accurate, timely, automated crop intelligence at scale.
Fortunately, aerial imagery from remote sensing, including satellite, airplane, and UAV, combined with image processing, computer vision, and more recently, deep learning, have led to numerous advances in precision agriculture to address these challenges.

Precision agriculture has seen rapid advancements in recent years due to the quality and availability of data combined with the advances in artificial intelligence and machine learning.
Many applications focus on ground-level imagery and point clouds to detect and classify disease and pests, precisely localize weeds from crop to enable automated spraying, reconstruct scenes to enable robotic harvesting, automate counting, and numerous other tasks to enable automation and improve efficiencies~\cite{grape_disease,wang2019review,3dreconmanipulate,corn_syngenta}.

Availability of more advanced remote sensing data has also made a tremendous impact on digital agriculture.
Key intelligence can be extracted at scale from imagery collected via satellite, plane, or UAV.
However, aerial images of any domain, pose many challenges for computer vision techniques as they tend to be massive in size, lower in resolution, and have statistics very different from those of natural images~\cite{statistics_1,marmanis2015deep,chiu2020agriculture}.
Imagery size is a particular challenge as high-resolution (\textless 1m/pixel) images are becoming more and more prevalent: a single field may be several GB in size and contain hundreds-of-millions pixels.
Approaches to improve receptive field size, multi-scale understanding, long-range interactions, and global context understanding in deep learning models has been an active area of research, although the focus has remained largely on natural images.
Some approaches include dilated convolutions, explicit multi-scale modeling, combined CNN and RNN modules, superpixels and graphical models~\cite{oord2016wavenet, Chen_2018_ECCV, chen2017rethinking, lin2017refinenet, Lin_2018_ECCV}.
The use of techniques specifically adapted to handle large images is important from the practical standpoint of inference speed and efficiency, in addition to performance improvements due to incorporating information from a larger portion of the image.

While pixel-level classification may be required in some scenarios, if the primary goal is to alert farmers to regions of the field exhibiting stress, classifying each pixel may be an unnecessarily fine-grained task.
Instead, we propose a method based on Simple Linear Iterative Clustering (SLIC) superpixels and a Graph Convolution Neural Network (GCN) to detect regions of nutrient deficiency stress (NDS) across a high-resolution image of field in an efficient manner.
We first over-segment the image into superpixels using SLIC and characterize each superpixel by a 9-element feature vector; these superpixels become the nodes of our GCN.
We then construct a fully-connected graph where each superpixel is connected to every other superpixel in the image, enabling information to propagate globally.
Initial weights of the edges are further determined by the similarities between the superpixels.
Finally, we compare two approaches in which the final task is either classification (whether any pixels in the super-pixel contained the target class) or regression (prediction of the fraction of target containing pixels directly).
Our approach effectively determines key areas of nutrient deficiency stress while using a network containing fewer than 5,000 parameters.

\section{Related Work}
\subsection{Computer Vision and Machine Learning for Precision Agriculture}

Precision agriculture is undergoing rapid transformation and advancement thanks to new sources of data and analytical techniques in computer vision.
The applications of machine learning in precision and digital agriculture are widespread including plant counting, weed detection, yield forecasting, pest and disease detection, storm damage quantification, high-throughput phenotyping, among others~\cite{malambo2019_sorghumcounting,barbosa2020modeling,sa2017weednet,boulent19_diseasecnn,derecho,phenotyping}.
Machine learning applied to these tasks is enabling farmers greater insight into the health of their crops, allowing them to make economically and environmentally better decisions in a more timely manner.  

The detection of nutrient deficiency stress is a key task for precision agriculture and has been explored extensively using traditional computer vision and deep learning approaches with both ground-level and remotely sensed images~\cite{mee2017detecting_ndsReview, wojtowicz2016application}. 
A large body of work relies on the use of hyperspectral imaging to identify stressed areas from changes in specific bands related to suppressed chlorophyll activity\cite{ wojtowicz2016application}.
Using machine learning approaches like support vector machines (SVMs) and artificial neural networks (ANNs), ~\cite{zha2020improving} was able to improve the estimation of nitrogen nutrient levels in rice from aerial imagery.

Recently, deep learning approaches have become increasingly common for machine vision applications in precision agriculture~\cite{kamilaris2018deep}.
Particularly relevant to this work, ~\cite{chiu2020agriculture} used a CNN approach to segment nutrient deficient areas, among other relevant agricultural patterns, from high resolution aerial imagery.   
Their work sought to determine NDS at the pixel-level and relied on a heavy, DeepLabV3-based architecture with several orders of magnitude more parameters than the GCN approach explored here.
Similarly, ~\cite{dadsetan2020detection} used a U-Net~\cite{unet} framework for detection and prediction of nutrient deficient areas; though lighter-weight than DeepLabV3, their U-Net is still significantly heavier than our proposed method.
Leveraging a subset of the dataset~\cite{chiu2020agriculture} which did not include the nutrient deficiency class, the authors of ~\cite{Liu_2020_CVPR_Workshops} proposed a Multiview Self-Constructing Graph Convolutional Network to segment the high-resolution images into six (plus background) classes. 
This approach used a CNN as a feature extractor followed by modules of self-construction graphs and graph convolutions before fusing to generate the final output and produced competitive results on the challenge dataset~\cite{Chiu_2020_CVPR_Workshops}. 

\subsection{Superpixels}
Superpixels are a common primitive in complex computer vision tasks because of their ability to produce a semantically meaningful and compact representation of images. 
The \textit{Simple Linear Iterative Clustering (SLIC)}~\cite{achanta2012slic} algorithm for producing superpixels has become one of the most prominent approaches for generating superpixels across a variety of domains including natural images\cite{gould2008multi}, medical images (e.g CT and mammography)~\cite{simarro2020leveraging} 
, and remote sensing~\cite{derksen2019scaling, zu2019slic} because of its speed and simplicity.
SLIC adapts k-means clustering to group pixels based on their similarity in color and proximity in the image.

Previously, superpixels have been combined with graphical models such as Conditional Random Fields (CRFs)~\cite{lafferty2001conditional} to enable the segmentation of images.
However, while these probabilistic graphical models successfully capture the spatial coherency of the image, they can be computationally burdensome as many methods rely on determining the maximum a posteriori (MAP) estimate to find the optimal label sequence.
This has led to an interest in combining superpixel methods with deep learning-based graphical methods, which will be explored in the next section.

Within the remote sensing domain, superpixels have been combined with a variety of machine learning approaches for numerous tasks.
To reduce computational time, \cite{superpixels_remote} applied a Random Forest classification based on SLIC superpixels to classify objects from remote sensing data with equivalent or superior accuracy and at a faster speed than pixel-based methods.
\cite{zhang2015superpixel} used superpixels with a graphical model for semantic segmentation of remote sensing image mapping.
Particularly relevant to the present work, ~\cite{kawamura2020discriminating} used a SLIC and Random Forest algorithm on images collected from UAVs to classify regions of a field as soil, crop, or weed.

\subsection{Graph Neural Networks}
Convolutional Neural Networks (CNNs) are commonplace in computer vision applications because their inductive bias efficiently captures the Euclidean structure of images~\cite{krizhevsky2012imagenet}.
Graph Neural Networks (GNNs), in contrast, offer a much greater degree of flexibility in the types of data which they can model~\cite{gnn_review, gori2005new, scarselli2008graph}.
This has led to their adoption in numerous domains including recommendation, molecular generation, and animation~\cite{bronstein2017geometric, ying2018graph, gilmer2017neural}.  

Graph Convolutional Networks (GCNs) are a further subset of GNNs which generalize the convolutional operator of CNNs over the domain of graphs and capture the dependencies between nodes.  
GCN approaches fall into one of two main categories: spectral~\cite{bruna2013spectral, kipf2017semisupervised} and spatial~\cite{duvenaud2015convolutional, atwood2015diffusion, hamilton2017inductive} approaches.  
Spectral methods are based on spectral graph theory and define the convolution operation in the Fourier domain by computing the eigendecomposition of the graph Laplacian~\cite{bruna2013spectral, defferrard2016convolutional}.
Spatial methods define the graph convolution operation based on a node's spatial relation in the graph as defined by its edges, and perform convolution by aggregating the neighboring nodes~\cite{monti2017geometric}.
Since the initial introduction of these methods, numerous advances have been made to incorporate attention, generative methods, adversarial methods, reinforcement learning, recurrent methods, and other structures common in the broader deep learning framework~\cite{velivckovic2017graph, zhang2020deep, yu2018learning, you2018graphrnn, 3dgcnn}.

GNNs have recently received increased attention from the remote sensing community.
To perform image classification on hyperspectral data, ~\cite{cai2020remote} used a cross-attention mechanism and GCN.
\cite{chaudhuri2019siamese} used a novel Siamese Graph Convolution Network (SGCN) to enable content-based image retrieval on remote sensing data.
Very relevant to the present work, \cite{ma2019attention} constructed an attention graph convolution network to perform image segmentation.  Their work leveraged single-channel, lower-resolution Synthetic Aperture Radar (SAR) as input and then generated superpixels before using a GCN to segment the regions of the image into different land usage categories.

\section{Methods}
\subsection{Data Collection and Preprocessing}
High-resolution (10cm/pixel) RGB aerial imagery of corn and soybean fields in Illinois, Indiana, and Iowa was collected during the 2019 growing season (April to October) using a Wide Area Multi-Spectral System (WAMS).
These images were mosaicked to create a single large (average 15k-by-15k pixels) image per field. 
Images underwent orthorectification using a digital elevation model of the field to produce a plainmetrically correct image~\cite{orthorect}.
Each field was imaged multiple times over the course of the season, but for this analysis, we focus only on the mid-season flights in which nutrient deficiency was likely to be present in the field.
A single image was selected from 317 fields during this period.
These were annotated by human experts for regions of nutrient deficiency stress; quality assurance (QA) was conducted after annotation.
The 317 flights were randomly separated into 223(70\%) train, 47(15\%) validation, and 47(15\%) test.

\subsection{Graph Generation from Superpixels}
\begin{figure*}[ht]
\begin{center}
   \includegraphics[width=1\linewidth]{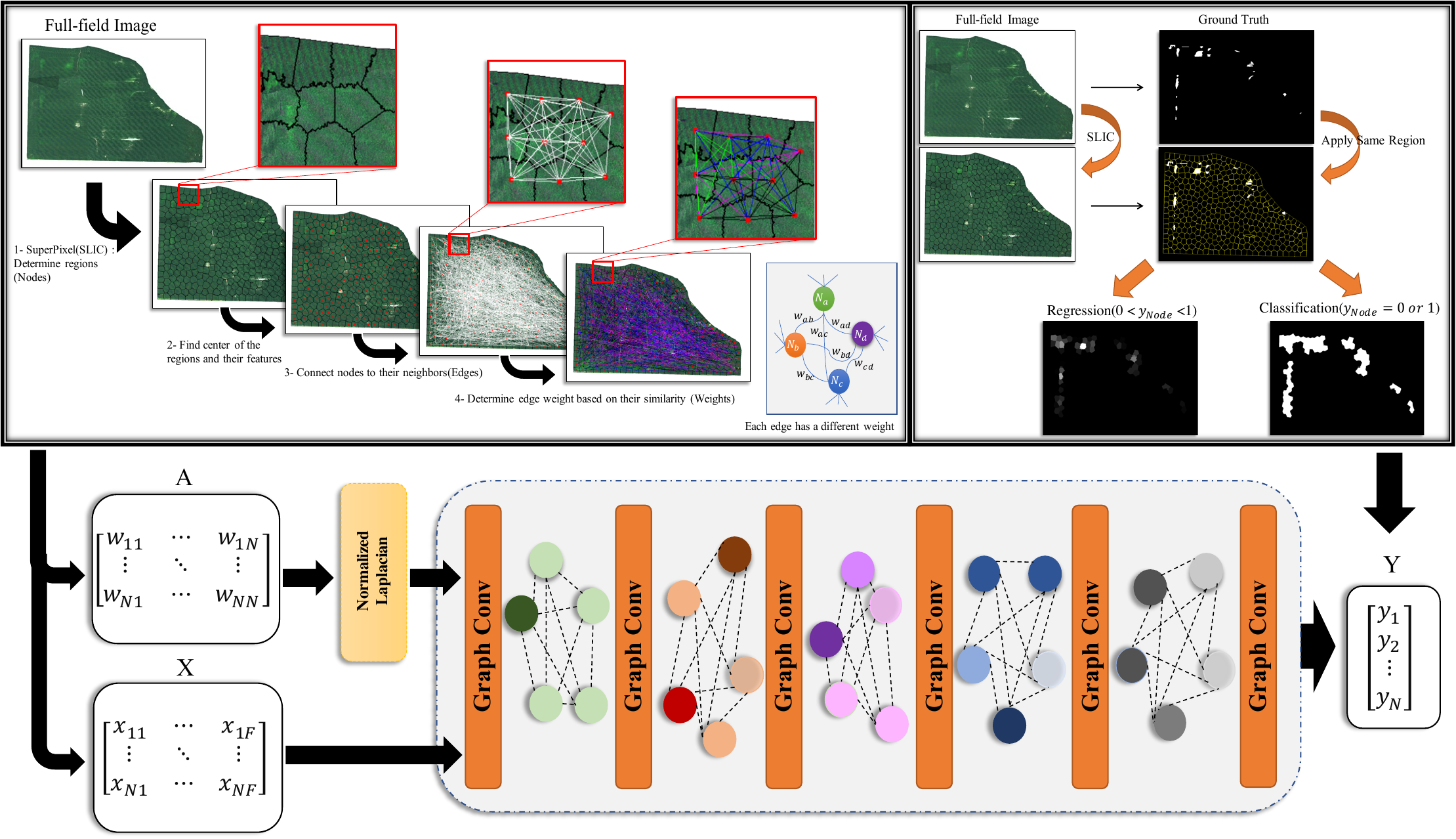}
\end{center}
   \caption{Our proposed method uses SLIC to first divide the image into superpixels.  
   Top Left: A fully connected graph is constructed over the superpixels. The nodes are defined by a 9-element feature vector derived from the RGB histogram within the superpixel and the initial edge weights are computed based on the RGB histogram similarity between pairs of superpixels. 
   Top Right: Node targets are generated for the respective regression and classification tasks.  For regression, the target node value is the fraction of pixels containing the positive class (i.e. NDS present) in the superpixel.  For classification, the node is labeled as 1 if \textit{any} pixel in the superpixel contained NDS and 0 otherwise.
   Bottom: The initialized adjacency matrix $\boldsymbol{A}$ and node feature matrix $\boldsymbol{X}$ are fed into six graph convolutional layers to predict the final node output $Y$.
   }
\label{fig:method}
\end{figure*}

Figure \ref{fig:method} shows an overview of our approach and is described by Algorithm~\ref{alg:graphgeneration}.

We use the SLIC algorithm to generate 400 segmented regions over the image, setting the initial compactness to 30. Because this algorithm uses k-means clustering, in some instances we need to add some neutral nodes which are only responsible to make the number of regions consistent at 400 across images. 

We define the undirected graph $\Graph$ as the set of nodes and edges given by 
$\Graph= \{\CHI, \mathcal{E}\}$
where
\begin{equation}
    \CHI = \{\boldsymbol{x_i} \in \mathbb{R}^F|i=1,...,N\}
\end{equation}
is the set of nodes with F-dimensional attributes and
\begin{equation}
    \mathcal{E} = \{\boldsymbol{e_{ij}} \in \mathbb{R}^S|\boldsymbol{x_i},\boldsymbol{x_j} \in \CHI\}
\end{equation}
is the set of edges with S-dimensional attributes.

Each segmented superpixel region $K_i$ becomes a single node in the subsequent graph; the region is described by the x and y coordinates of the superpixel centroid and the multidimensional RGB color histogram ($H_i$) in that region. 
As these are narrow-band images, many of the elements in a given channel are zero, making the distribution strongly non-Gaussian. 
Therefore we constructed an adjusted histogram, looking at only the non-zero elements to construct our node-feature vector.
From this adjusted histogram we constructed $F=9$ initial features: the three-element mean $\boldsymbol{\mu_i}$, the three-element standard deviation $\boldsymbol{\sigma_i}$, and the three-element fractional activation $\boldsymbol{\alpha_i}$.
This last feature captures the fraction of elements in each of the original channels that are non-zero.

We connect all ($N=400$) nodes to generate a fully-connected graph and initialize the edge weights as follows.
First, we compute the Bhattacharyya coefficient\cite{bhattacharyya1943measure} $BC_{ij}$ between the multi-dimensional RGB histograms, $H_i$ and $H_j$, of pairs of nodes $K_i$ and $K_j$.

We define the similarity between nodes as $W_{ij}=1-BC_{ij}$.
As the Bhattacharyya coefficient is defined on [0,1], this maps nodes with low overlap (i.e. high BC value) to a similarity score close to 0 and nodes with high overlap (i.e. low BC value) to a similarity score close to 1. 
We then set our initial edge weight to this similarity measure.

With this initialization, we represent $\Graph$ by its adjacency matrix $\boldsymbol{A}\in\{0,1\}^{N \times N}$, node features $\boldsymbol{X}\in{\mathbb{R}^{N\times F}}$, and edge features $\boldsymbol{E}\in{\mathbb{R}^{N\times N\times S}}$.
Applying the \textit{renormalization trick} according to \cite{kipf2017semisupervised}, we define the convolved signal matrix $Z \in \mathbb{R}^{N\times F}$ as
$Z = \tilde{D}^{-\frac{1}{2}}\tilde{A}\tilde{D}^{-\frac{1}{2}}X\theta$
where $\theta\in\mathbb{R}^{C\times F}$ is the matrix of filter parameters for C input channels and
$\tilde{D}_{ii} = \sum_{j}\tilde{A_{ij}}$ is a layer-specific trainable matrix of weights.

\subsection{Target Map Creation}
We explore two approaches(classification and regression) using this framework and construct an appropriate target map for each as seen in the top-right sub-box of Figure~\ref{fig:method}.

For the ``classification'' target, the node's label will be binary. We define the node's label $y_i = 1$ if \textit{any} ground-truth pixel with label 1 is inside that node's region, otherwise, it is considered $0$. 
This has the effect of boosting the presence of the NDS class, which is rare.
A different threshold could be used to generate the classification target which may make the task easier for the model to learn, however, we chose this approach because of the desire for the model to aggressively identify regions of stress.

In contrast, for the ``regression'' target, we seek to identify the fraction of pixels within a superpixel that contain NDS.
We determine the label of each node $y_i$ as a proportion of pixels with label 1 (i.e. NDS present) inside each superpixel region boundary. 
In this way, a number between 0 and 1 is assigned as the label of that node.

\begin{algorithm}
\DontPrintSemicolon
\caption{Graph Generation from superpixel}
\label{alg:graphgeneration}
\KwData{Image $I$, Superpixel technique $SLIC$, Number of nodes $N$, Task $T$}
\KwResult{Adjacency Matrix $\mathcal{A}$ , Feature Matrix $\mathcal{X}$, Label Vector $\mathcal{Y}$}
\Begin{ \tcp*{Adjacency \& Feature Matrix Creation}
  $ Nodes:\{K_1,..K_N\} \longleftarrow SLIC(I, N)$\\
 \For{$K_i$ in $Nodes$}{
     $ \mathcal{X}_i \longleftarrow $Feature-Extractor$(K_i)$
   
     $H_i \longleftarrow$ MultiD-Histogram$(K_i)$
 }
 \For{$K_i$ in $Nodes$}{
    \For{$K_j$ in $Nodes$}{
         $BC_{ij} \longleftarrow $ Bhattacharyya-Coef$(H_i,H_j)$
         $\mathcal{A}_{ij} \longleftarrow 1-BC_{ij}$
    }
 }
 }
\Begin{ \tcp*{Target Map Creation}
 
 \If{$T$ is $``Classification"$} { 
     $\mathcal{Y}_i \gets 1 \quad $if in $K_i$ any pixels contain the target class $1$ 
    }
 
 \If{$T$ is $``Regression"$} {
     $\mathcal{Y}_i \gets$ fraction of target class $1$ containing pixels in $K_i$ 
    }

}
\end{algorithm}

\subsection{Model Architecture and Training}
The above adjacency matrix $\boldsymbol{A}$ and node feature matrix $\boldsymbol{X}$ are input into our neural network.
We use six graph convolutional layers with [320, 1056, 1056, 1056, 1056, 33] parameters in each respective layer. The first five layers have $l2$ kernel regularization and $elu$ activation. The last graph convolutional layer uses a $sigmoid$ activation function to produce the output $\boldsymbol{Y}$.

For both tasks, the loss is computed as the Dice Loss~\cite{dice} between the target and predicted maps.
We elect to use Dice loss for the regression task because the value being regressed is a value between 0 and 1 corresponding to the fraction of pixels in the superpixel containing the positive class.
The matrix multiplication to compute the intersection of the target and predicted maps is computed as in the classification tasks except that the predicted map is continuous instead of binary.
This has the effect of ``softening'' the impact from nodes which have minimal NDS.

All models are trained with a batch size of 32 for 200 epochs using Adam Optimization with an initial learning rate of 1e-3.
This value is decayed by a factor of 10 if no improvement to the validation loss is seen for a period of 10 epochs.

The Graph Convolutional Network is build in Spektral~\cite{grattarola2020graph} and Keras (version 2.2.5) with a Tensorflow (version 1.15) backend.
Models are trained on a machine with 1 NVIDIA Tesla V100 GPUs with 32GiB memory in total.

\section{Results}

\begin{table*}
\begin{center}
\begin{tabular}{llllll}
\hline
\multicolumn{1}{c}{Model}               & Dice Loss & Precision & Recall & F1-score & IOU-score \\ \hline
Graph convolution- classification  &0.83 & 0.33      & 0.65   & 0.43 & 0.28\\
Graph convolution- regression   & 0.95      & *   & * & * & *    
\\ \hline
\end{tabular}
\end{center}
\caption{Results from the GCN-based classification and regression models on the test set.}
\label{tab:results}
\end{table*}

\subsection{Classification Model}
Results of the classification model are shown in Table~\ref{tab:results}.
This model had a dice loss of 0.83.
We then threshold the node probabilities at $p = 0.4$ to compare the regional overlap with the target map; at this threshold the precision is $0.33$, recall is $0.65$, F1-score is $0.43$ and IOU-score is $0.28$. Note that the threshold is adjusted to 0.4 based on precision-recall curve to achieve the highest F1-score. 
Recall that we set the target value of the node to 1 if any pixel in that superpixel contained NDS; this is an extremely difficult task as the model may be penalized for miss-classifying a node even if it has an very minimal amount of NDS.
Exploring this thresholding is the focus of future work.

\begin{figure*}[ht]
\begin{center}
   \includegraphics[width=0.7\linewidth]{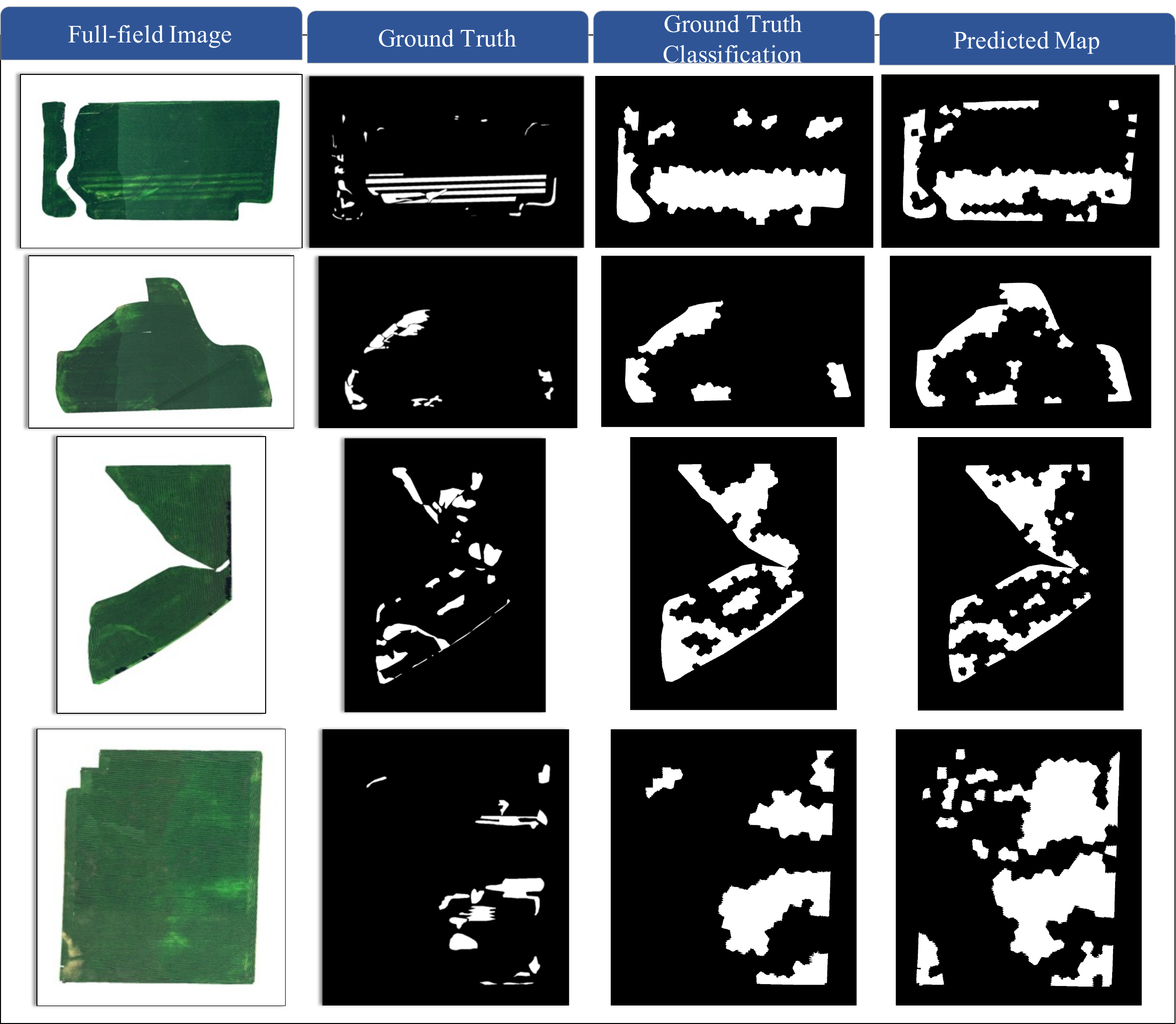}
\end{center}
   \caption{Example results for the classification model. The Ground Truth map shows pixel-level annotations and while the Ground Truth Classification and Predicted Map show node-level labels at the superpixel level.}
\label{fig:Classification_results}
\end{figure*}

Qualitative results are shown in Figure~\ref{fig:Classification_results}.
We see that the model does a good job identifying key areas of nutrient deficiency stress.
The first row shows a clear example of how the use of superpixels and the GCN finds the large region near the bottom of the field exhibiting NDS; while a more detailed pattern exists at the pixel level, those fine-grained details are not necessary to alert the farmer to this key problem in the field.

\subsection{Regression Model}
In creating the target nodes for the classification task, we made the decision to label the node as a positive class if any pixel in the superpixel contained NDS; this is inevitably removing information about the extent of NDS in that region.
Therefore, we sought to predict the fraction of pixels labeled as nutrient deficient directly through a regression task.

\begin{figure*}[ht]
\begin{center}
   \includegraphics[width=0.7\linewidth]{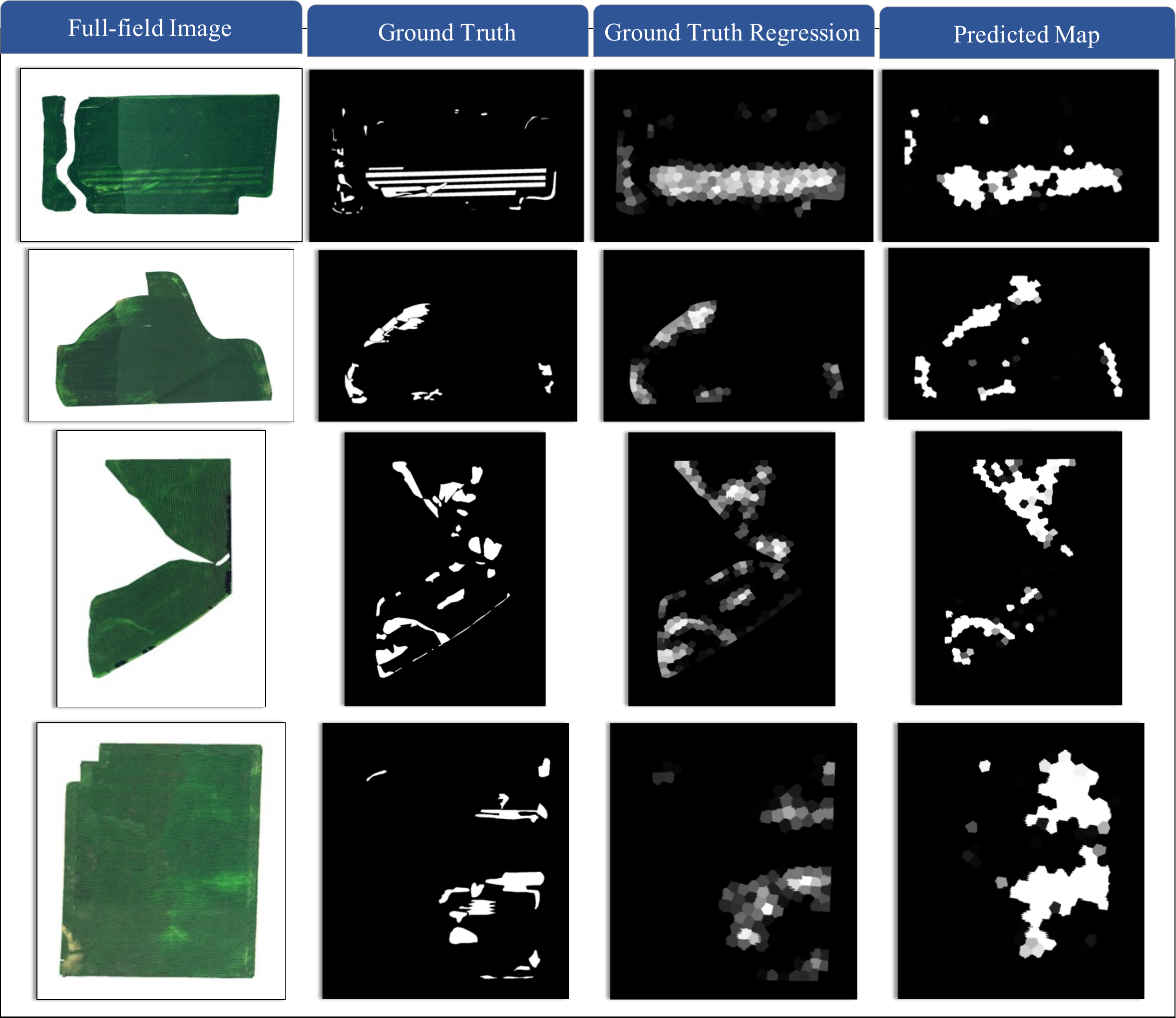}
\end{center}
   \caption{Example results for the regression model. The Ground Truth map shows pixel-level annotations and while the Ground Truth Regression and Predicted Map show node-level labels at the superpixel level.}
\label{fig:regression_results}
\end{figure*}

The Dice loss of the regression model is 0.95 as seen in Table~\ref{tab:results}. 
The predictions obtained via direct regression of the fraction of NDS in a superpixel are shown in \ref{fig:regression_results}.
These results demonstrate that regression model did not perform as well as the classification task even though a continuous target was provided.
We elected to use Dice loss of both tasks for more direct comparison, however, other regression-appropriate loss functions, like RMSE, could be used instead and may improve performance; this will be explored in future work.

\section{Discussion}

\subsection{Performance and Comparison to Other Methods}
Both the classification and regression models identify key areas of nutrient deficiency stress from high-resolution RGB aerial imagery.
We used an ``aggressive'' target creation step for the classification model, setting a node with any pixels containing NDS to 1.
This threshold could be tuned as a hyperparameter to alter the model's output depending on the desired balance between precision and recall; a value which produces the highest F1 score may be chosen, alternately there may be reasons to value higher precision at the expense of a lower recall.
Although the regression model did not perform as well, we believe further exploration around different loss functions may lead to enhanced performance.

Detection of nutrient deficiency stress from remotely sensed data is not a new tasks with past approaches spanning traditional image processing as well as more modern deep learning methods.
Recent deep learning approaches have performed well on the task, although they have been largely focused on pixel-level classification.
The DeepLabv3+ multi-class models of \cite{chiu2020agriculture} demonstrated an IOU on the nutrient deficiency class in the range 0.3418 - 0.3940 (validation) and 0.3937 - 0.4415 (test) on their dataset.
In \cite{dadsetan2020detection}, the authors achieved a top single-image IOU of 0.34 and F1 Score of 0.43 using a U-Net with EfficientNet backbone.
Because our classification is at the node (i.e. superpixel) level and uses only RGB (no NIR channel), it is not directly comparable to these approaches.
We believe that incorporating additional channels and vegetative indices within the node features will further improve the models framework.

Additionally, our model with only 4,577 parameters is orders of magnitude smaller than either of these approaches, each which has tens-of-millions of parameters.
Both the classification and regression models trained in a matter of minutes: over 200 epochs could be run on readily available hardware in around 5 minutes and GCN inference can run at ~650frames/sec.
Pixel-level accuracy may be a requirement for certain tasks, however, in the regime in which identification of regions is sufficient, our approach offers tremendous advantages on the computational front.
As the amount of data grows through additional channels (e.g. hyperspectral imagery) or revisits (i.e. longitudinal imagery) this can become further computationally burdensome to process.
As the goal of such a model is to provide farmers with intelligence about their fields in order to improve economic efficiencies while minimizing any negative environmental impact, such considerations around model design are important.

\subsection{Future Work}
For our node features, we focus only on 9 simple features which characterized the RGB histogram.
Future work will explore the use of more complex features including additional moments of the distribution as well as additional channels (like NIR) and indices (like NDVI, GNDVI). 
The GCN framework enables the addition of other relevant features such as these without dramatically increasing the size or complexity of the network.

Similarly, we have chosen to focus on only a single (binary) task for this analysis.
However, this method can easily be extend to other patterns of interest such as flooding, inhibited emergence, drydown, weeds, etc..  

Furthermore we have focus only on a single point in time for this analysis. 
~\cite{dadsetan2020detection} demonstrated the significant boost in performance which can be achieved through the use of longitudinal data.
The superpixel-GCN framework is well posed to handle longitudinal data.
A key advantage of this framework is the ability to handle registration shifts which may occur across multiple images.
More specifically, perfect pixel-level alignment is not guaranteed across the sequence of images.
While CNN-based approaches have been successful even despite this noise, the GCN approach offers additional flexibility as the graph can be constructed to enable information flow across all regions of all images.

Finally, in this work we have used a very simple GCN to demonstrate the usefulness and effectiveness of this approach.
Additional modules such as attention~\cite{velivckovic2017graph} have proven to be highly successful in both natural image and remote sensing domains~\cite{cai2020remote} and are the focus of future work.

\subsection{Conclusion}
In this work we have presented an approach with uses a Graph Convolutional Neural Network on top of SLIC Superpixels to efficiently detect regions of nutrient deficiency stress.  
Our model identifies regions of the field experiencing nutrient deficiency stress and has only 4,577 parameters.
This work has only used a limited set of node features and uses only RGB imagery, however, we believe the performance of this approach demonstrates its usefulness and viability to detect NDS and other relevant agricultural patterns from large remotely sensed imagery in a computationally efficient manner. 



{\small
\bibliographystyle{ieee_fullname}
\bibliography{main}
}
\end{document}